\documentclass[11pt]{article}
\usepackage{times}
\usepackage{fullpage}
\usepackage{adjustbox}
\usepackage[most]{tcolorbox}
\usepackage{comment}
\usepackage{placeins}
\usepackage{caption}

\usepackage[numbers]{natbib}
\usepackage{graphicx}

\usepackage{url}

\title{Discovering Software Parallelization Points Using Deep Neural Networks}

\author{
  Izavan dos S. Correia \\
  Federal Rural University of Pernambuco \\
  Recife, Brazil \\
  \texttt{izavan.correia@ufrpe.br}
  \and
  Henrique C. T. Santos \\
  Federal Institute of Pernambuco \\
  Recife, Brazil \\
  \texttt{henrique.santos@recife.ifpe.edu.br}
  \and
  Tiago A. E. Ferreira \\
  Federal Rural University of Pernambuco \\
  Recife, Brazil \\
  \texttt{tiago.espinola@ufrpe.br}
}

\date{2025}

\begin{document}

\maketitle

\begin{abstract}
This study proposes a deep learning-based approach for discovering loops in programming code according to their potential for parallelization. Two genetic algorithm-based code generators were developed to produce two distinct types of code: (i) independent loops, which are parallelizable, and (ii) ambiguous loops, whose dependencies are unclear, making them impossible to define if the loop is parallelizable or not. The generated code snippets were tokenized and preprocessed to ensure a robust dataset. Two deep learning models — a Deep Neural Network (DNN) and a Convolutional Neural Network (CNN) — were implemented to perform the classification. Based on 30 independent runs, a robust statistical analysis was employed to verify the expected performance of both models, DNN and CNN.
The CNN showed a slightly higher mean performance, but the two models had a similar variability. Experiments with varying dataset sizes highlighted the importance of data diversity for model performance. These results demonstrate the feasibility of using deep learning to automate the identification of parallelizable structures in code, offering a promising tool for software optimization and performance improvement.
\end{abstract}

\textbf{Keywords:} Software Engineering, Machine Learning, Program Analysis, Loop Classification, Parallelization Detection, Deep Neural Networks, Convolutional Neural Networks, Genetic Algorithm, Code Tokenization

\section{Introduction}
\label{sec:introduction}

In today's rapidly evolving technological landscape, optimizing software performance has become more crucial than ever. One of the most effective strategies for achieving this optimization is parallel programming. By harnessing the power of parallel computing, developers can significantly enhance the speed and efficiency of their applications\citep{Santos2025}.

Parallel programming allows tasks to be divided into smaller, independent units that can be executed simultaneously across multiple processing cores or even distributed computing systems. This approach not only reduces execution time but also maximizes resource utilization, resulting in faster and more responsive software.

As a result, the growing demand for faster computing has made parallel programming essential in modern software development. However, identifying code segments that can run in parallel remains a challenge, especially when dealing with legacy systems or code written by others\citep{garcia2010bridging}. This issue resembles the challenge of debugging, where subtle structural details can greatly impact performance\citep{di2018detecting}. Current methods, whether manual\citep{wolfe1995high} or tool-assisted\citep{garcia2010bridging}, struggle with: (1) identifying implicit loop dependencies, and (2) scaling to modern codebases.

Machine learning has introduced new methods for analyzing code, demonstrating strong capabilities in pattern recognition and classification\citep{allamanis2018survey}. Notable examples include the detection of code quality issues\citep{Sharma_2021} and enhancements in software design\citep{das2019detecting}, which closely relate to identifying parallelizable code. This study aims to develop and evaluate an automated system that accurately identifies loops capable of being executed in parallel across various types of code.

One of the most common and effective opportunities for parallelization in software lies in loop structures. When the iterations of a loop are independent—meaning each iteration does not rely on the result of another—they can be executed in parallel rather than sequentially. This makes loops ideal candidates for parallel programming. Accordingly, the core idea of this work is to apply deep neural networks to detect when a loop can be parallelized.

To apply the deep learning algorithms, a dataset of codes was created. We employ a genetic algorithm to generate various code examples, including both parallelizable and non-parallelizable codes, framing the problem as a binary classification task. This approach enables the construction of a training dataset that addresses the need for diversity, as highlighted in prior research\citep{liu2019deep}. Next, we evaluate two types of neural networks: Deep Neural Networks (DNNs), which are effective in learning complex code patterns\citep{goodfellow2016deep}, and Convolutional Neural Networks (CNNs), originally developed for image analysis\citep{lecun1998gradient} but adapted here for code analysis, following their application in other non-visual domains\citep{sun2020city}.

Traditional methods for detecting parallelizable code can be broadly categorized into two approaches. The first involves static code analysis, which uses techniques such as control and data flow analysis to identify dependencies that prevent parallel execution\citep{wolfe1995high}. While effective for simple and well-structured programs, these methods often struggle with complex or obfuscated code and may miss implicit dependencies.

The second category includes reinforcement learning (RL)-based techniques, where agents learn to annotate or transform code by interacting with an environment and receiving feedback in the form of rewards\citep{cummins2017end, colbert2021generating}. These approaches are promising and have shown performance gains—for example, Cummins et al.\citep{cummins2017end} report improvements of up to 14\% in compiler heuristics. However, they require large datasets, expert-crafted features, prolonged training times, and integration with real compiler pipelines, making them resource-intensive and less accessible for rapid prototyping. Colbert et al.Colbert et al.\citep{colbert2021generating}, for instance, rely on long benchmark executions and complex tuning procedures for GPU optimization using off-policy RL.

In contrast, our approach leverages deep learning models to directly classify loops based on learned structural patterns, offering potential advantages in speed, scalability, and automation. Our models are trained on approximately 4,000 synthetically generated code samples using a genetic algorithm, allowing for broad coverage of parallelizable and non-parallelizable loop structures. Training is completed in a matter of hours on a single GPU, significantly reducing resource requirements compared to RL-based approaches. We also apply Principal Component Analysis (PCA) to reduce input dimensionality\citep{pearson1901liii, Gray2017}, helping to improve generalization and model performance, as supported by recent studies\citep{khleel2022deep}. Our results indicate high classification accuracy and suggest that deep learning may provide meaningful improvements over traditional techniques, particularly for the automated identification of parallelism opportunities.

The remainder of this paper is organized as follows: Section 2 presents the proposed methodology, Section 3 analyzes the experimental results, and Section 4 concludes the paper and outlines future work directions.

\section{METHODOLOGY}
\label{sec:methodology}

\subsection{Programming Code Generation}

To support model training, we implemented a code generator based on a genetic algorithm using the \texttt{DEAP} Python library\footnote{DEAP: Distributed Evolutionary Algorithms in Python. Available at: \url{https://deap.readthedocs.io}}. Python was used as the programming language for both the generator and the generated examples; however, the approach can also be applied to other languages.

Two separate code generators were implemented: one for producing parallelizable examples (positive class) and another for generating non-parallelizable examples (negative class). Both generators share the same underlying logic and structure, differing only in how loops are constructed. In the positive class, loop variables are not reused within the loop body, ensuring independence; in the negative class, variables are deliberately reused or coupled with dependent assignments, simulating ambiguous or non-parallelizable scenarios.

Each sample includes common programming constructs such as function definitions, variables, operations, conditionals, libraries, and loops. The primary distinction between the two classes lies in the loop structure, as described above.

A genetic algorithm was used to evolve syntactically valid code samples. Each individual in the population encodes a complete Python program as a string. The initial population is randomly generated and then evolves through genetic operations. Crossover is performed by exchanging code line segments between individuals, whereas mutation applies small edits, such as variable modifications or the insertion of print statements.

The fitness function rewards structural features such as the number of functions, the presence of loops and conditionals, variable usage, and successful compilation. Invalid or overly simplistic programs are penalized. Formally, the fitness score $f$ for a given individual program $P$ is defined as:

\begin{equation}
\label{eq:fitness}
f(P) = \sum_{i=1}^{7} s_i + s_{\text{comp}}
\end{equation}

Where each $s_i$ is a score assigned to a specific structural feature of $P$, and $s_{\text{comp}}$ is a compilation score that strongly penalizes programs that fail to compile. Specifically, each score is assigned as follows:

\begin{itemize}
\item $s_1$: number of \texttt{import} statements is between 1 and 12 $\Rightarrow +1$, otherwise $-1$;
\item $s_2$: number of \texttt{def} functions is between 2 and 8 $\Rightarrow +1$, otherwise $-1$;
\item $s_3$: number of \texttt{if} conditionals is between 2 and 8 $\Rightarrow +1$, otherwise $-1$;
\item $s_4$: number of \texttt{print} calls is between 2 and 8 $\Rightarrow +1$, otherwise $-1$;
\item $s_5$: number of variables is between 2 and 100 $\Rightarrow +1$, otherwise $-1$;
\item $s_6$: number of \texttt{for} loops is between 1 and 6 $\Rightarrow +1$, otherwise $-5$;
\item $s_7$: number of lines is between 10 and 150 $\Rightarrow +1$, otherwise $-1$;
\item $s_{\text{comp}}$: if the program compiles successfully $\Rightarrow +1$, otherwise $-5$.
\end{itemize}

The evolution process uses a population of 10{,}000 individuals over 50 generations, with roulette selection, 0.9 crossover probability, and 0.1 mutation probability.

\begin{figure}[!ht]
\centering
\includegraphics[width=\linewidth, height=6cm, keepaspectratio]{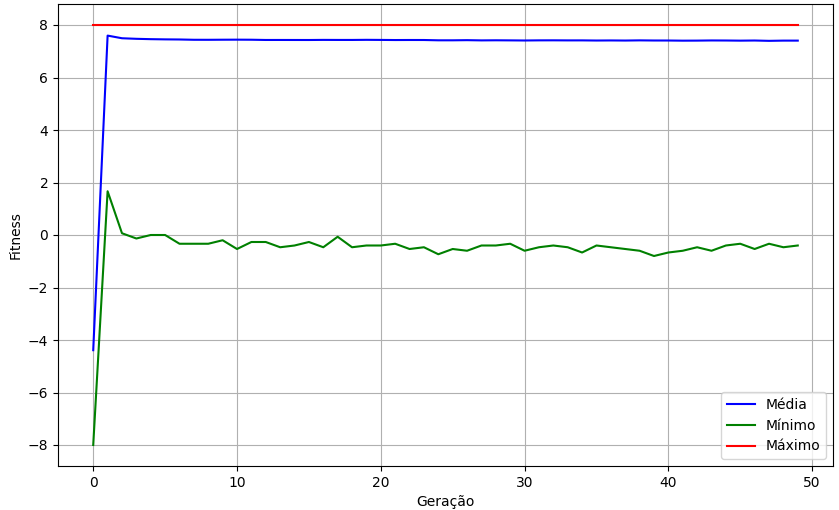}
\caption{Fitness evolution across 30 runs of the genetic algorithm. The plot shows the average, maximum, and minimum fitness per generation.}
\label{fig:fitness_curve}
\end{figure}

To demonstrate the optimization progress and the quality of the generated solutions, Figure ~\ref{fig:fitness_curve} shows the average, maximum, and minimum fitness values across 30 independent runs of the genetic algorithm. The curves indicate consistent improvement in population fitness early on, even with just 50 generations. Over time, fitness remains steady. For this study, 50 generations were sufficient to generate diverse and well-structured code samples aligned with the defined fitness criteria.

In addition to tracking optimization progress, we used the Hall of Fame mechanism provided by the \texttt{DEAP} framework, which stores the best individuals found during the evolutionary process, to retain the 500 best individuals from a total population of 10,000 candidates. All selected individuals were validated in the VSCode\footnote{Visual Studio Code editor, developed by Microsoft. Available at: \url{https://code.visualstudio.com/}}
 environment and did not present any compilation issues, which reinforces the robustness of the evolutionary approach and the quality of the generated solutions.
 
\begin{figure}[!hbt]
    \centering
    \vspace{-0.5em}
    \includegraphics[width=\linewidth, height=4cm, keepaspectratio]{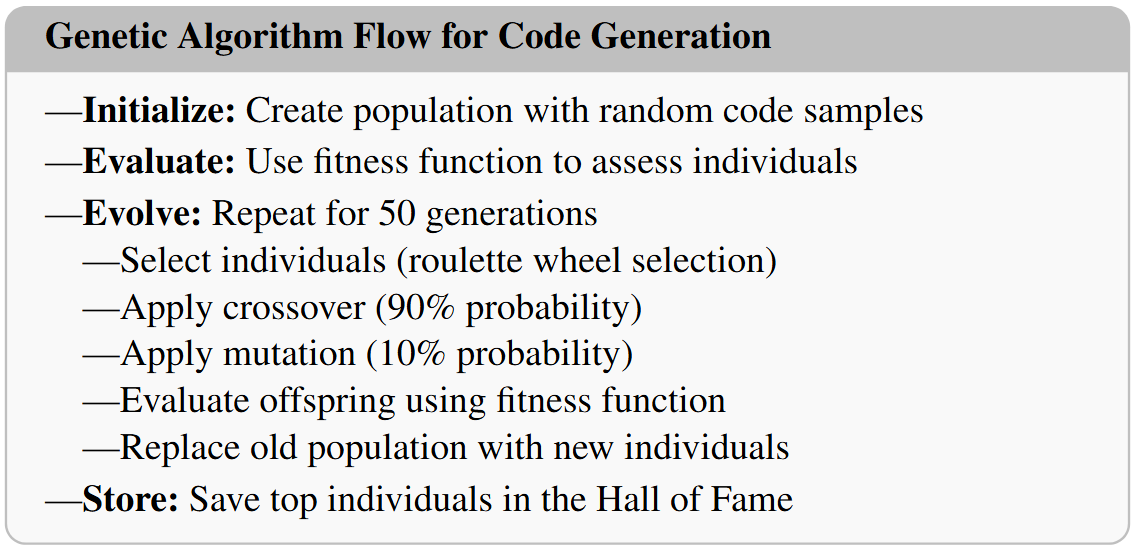}
    \vspace{-0.6em}
    \caption{Evolutionary process for generating labeled code samples.}
    \label{fig:ga-flow}
\end{figure}

This process guarantees the generation of diverse and realistic code samples for both classes while strictly adhering to the structural constraints imposed by the fitness function. A general overview of the entire genetic programming pipeline, including key stages such as initialization, selection, crossover, and mutation, is illustrated in Figure~\ref{fig:ga-flow}.

\subsection{Dataset Creation}

We generated a dataset of 4000 samples using the code generator:
\begin{itemize}
    \item 2000 labeled as parallelizable (independent loops).
    \item 2000 labeled as non-parallelizable (dependent or ambiguous loops).
\end{itemize}

To build this dataset, the generator for independent codes was executed four times, producing the 2000 samples of the first class. Similarly, the generator for non-parallelizable codes was executed four times to produce the 2000 samples of the second class, resulting in a total of eight generator executions.

After generation, all code samples were tokenized using Python’s \texttt{tokenize} library\footnote{\url{https://docs.python.org/3/library/tokenize.html}}. Each token — such as keywords, identifiers, or operators — was mapped to a unique numerical identifier and stored in a JSON dictionary to ensure consistent mapping. Identifiers were reused for repeated tokens. For example, the token \texttt{"import": 1} indicates that the keyword \texttt{import} was assigned to the identifier \texttt{1}. 

This process yielded a set of integer sequences, each labeled as either class 1 (parallelizable) or class 0 (non-parallelizable).

\subsection{Data Processing and Preparation}

The tokenized sequences were used as input to the neural networks. To reduce dimensionality, we applied Principal Component Analysis (PCA). Models were trained using:
\begin{itemize}
    \item The full original data (retaining 100\% variance).
    \item Reduced representations preserving 95\%, 90\%, 85\%, and 80\% of variance.
\end{itemize}

We normalized the input data and split it into training (70\%), validation (15\%), and testing (15\%) sets.

\subsection{DNN Model Architecture}

We implemented the deep neural network (DNN) using \texttt{PyTorch}\footnote{\url{https://pytorch.org/docs/stable/index.html}}.  
Preliminary experiments were conducted to explore different architectural configurations, and based on the results, the selected architecture was identified as the most promising for the task at hand.  
The architecture consisted of the following:

\begin{itemize}
    \item First layer: 128 units, followed by Batch Normalization, ReLU, and Dropout (0.5).
    \item Second layer: 64 units, Batch Normalization, ReLU, and Dropout (0.5).
    \item Third layer: 32 units, Batch Normalization, ReLU, and Dropout (0.5).
    \item Output layer: 1 unit with sigmoid activation (binary classification).
\end{itemize}

We applied Batch Normalization after each linear layer, followed by ReLU activation and Dropout. The model was trained for 1000 epochs using Binary Cross-Entropy Loss (\textit{i.e.}, \texttt{BCELoss}\footnote{\url{https://pytorch.org/docs/stable/generated/torch.nn.BCELoss.html}}) and the Adam optimizer~\cite{kingma2014adam} with a learning rate of 0.001 and batch size of 4.

We logged training metrics (accuracy and loss) every epoch. After training, we evaluated the model’s performance on the test set using accuracy, binary cross-entropy loss, a confusion matrix, and a classification report (precision, recall, F1-score).

\subsection{CNN Model Architecture}

We also implemented a convolutional neural network (CNN) using \texttt{PyTorch},  
following the same approach as for the DNN and adopting the configuration that proved most promising in our preliminary tests. The architecture consisted of the following:

\begin{itemize}
    \item First convolutional layer: 2 filters, kernel size 3, stride 1, padding 1; followed by Batch Normalization, ReLU, and Dropout (0.6).
    \item Second convolutional layer: 4 filters, kernel size 3, stride 1, padding 1; followed by Batch Normalization, ReLU, and Dropout (0.6).
    \item Fully connected layer: 4 units with ReLU activation.
    \item Output layer: 1 unit with sigmoid activation.
\end{itemize}

The output feature maps from the convolutional layers were flattened and fed into the fully connected layers. The same training setup used for the DNN—1000 epochs, Adam optimizer, BCELoss, learning rate of 0.001, and batch size of 4—was used for the CNN.

We evaluated performance using accuracy, loss, confusion matrix, and a classification report.

\subsection{Training and Evaluation}

Each model was trained and evaluated over 30 independent runs to account for variability due to random initialization and data shuffling. We trained the models on the training set and validated them after each epoch. Evaluation was conducted on the testing set by comparing predicted and true labels.

Model performance was assessed using:
\begin{itemize}
    \item Accuracy and binary cross-entropy loss.
    \item Confusion matrix.
    \item Precision, recall, and F1-score based on a classification report.
\end{itemize}

Additionally, we generated training history plots to illustrate the progression of loss and accuracy across epochs.

\section{Results}
\label{sec:results} 

This section presents the results obtained from experiments conducted with both the DNN and CNN models, based on 30 executions for each model to ensure statistical reliability and mitigate the effects of random initialization and data shuffling. The evaluation metrics include accuracy, precision, recall, F1-score, and Binary Cross-Entropy Loss. 

The presentation of results follows a progressive analytical structure: first, we provide an overview of the 30 executions using boxplots to illustrate the performance distribution of each model. Then, we analyze the average behavior of the DNN and CNN models across these executions, highlighting key differences. Next, we incorporate a Principal Component Analysis (PCA) to visualize how performance varies across different training set proportions, followed by an examination of the models' average behavior at each of these proportions. Finally, we identify and analyze the best and worst executions for each model, offering insights into their variability and robustness.

\subsection{Performance Distribution Across 30 Executions}

\begin{figure}[!htb]
\centerline{\includegraphics[width=\linewidth, height=4.5cm, keepaspectratio]
{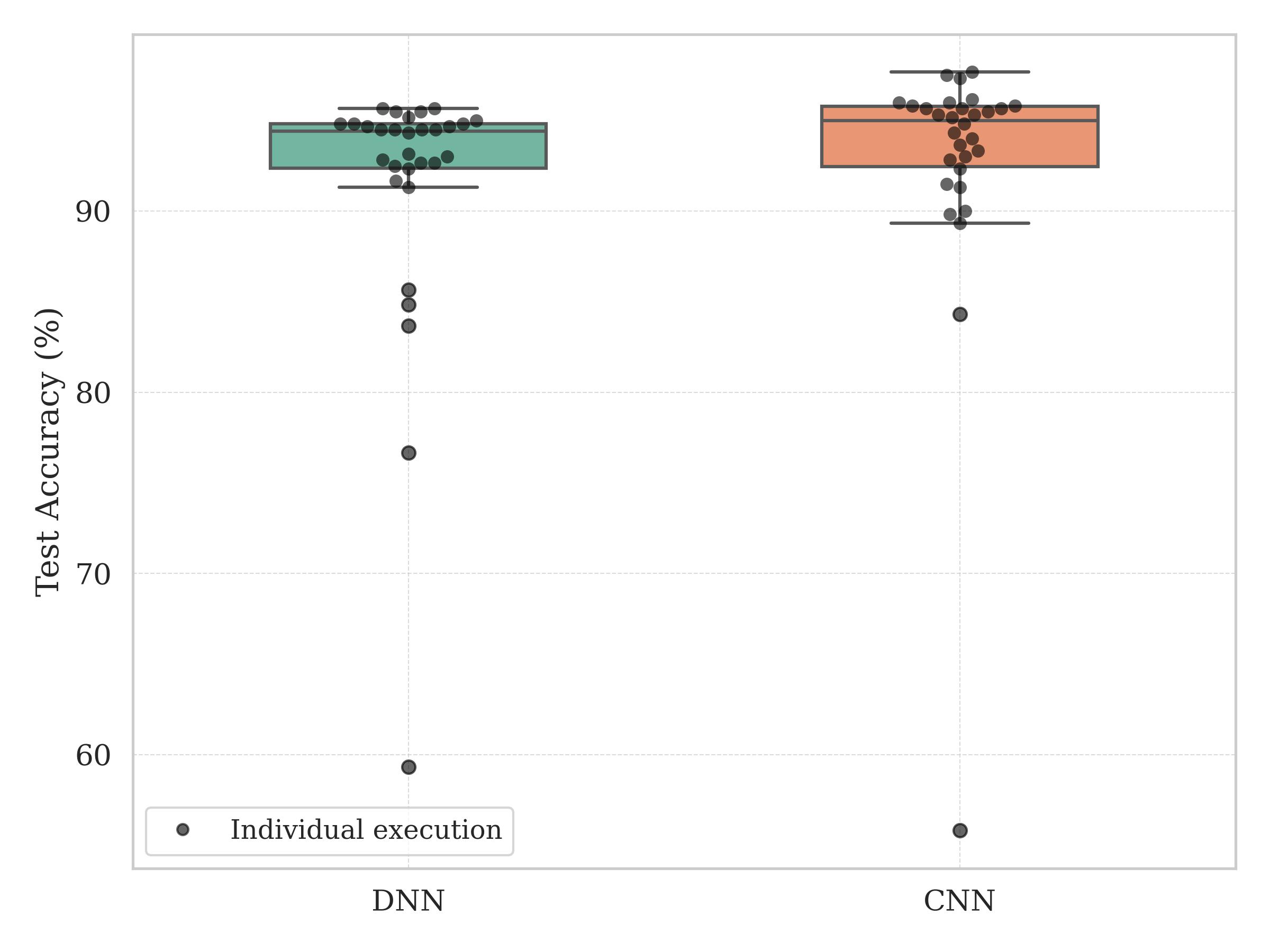}}
\centerline{\includegraphics[width=\linewidth, height=4.5cm, keepaspectratio]{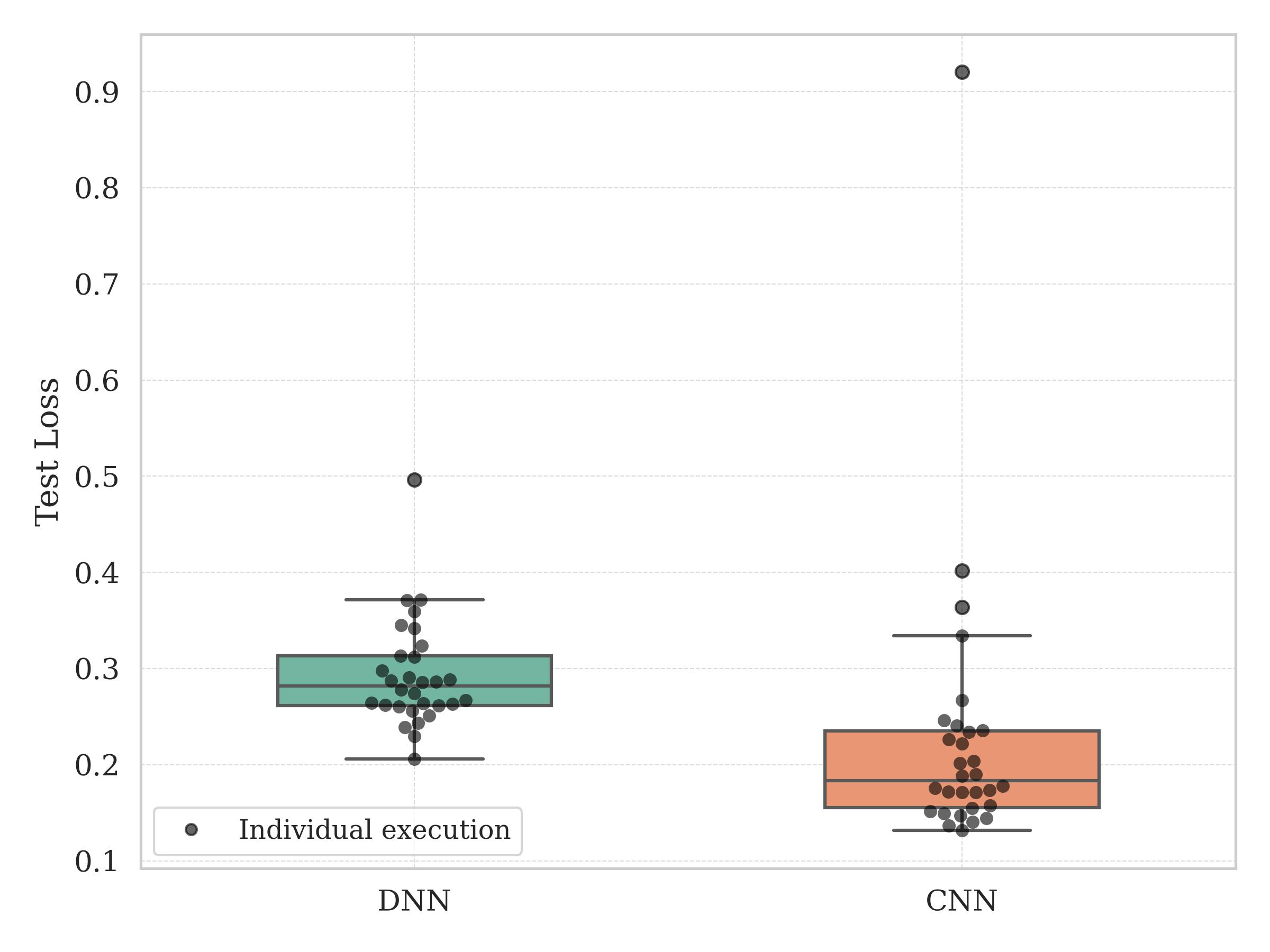}}
\caption{Boxplots showing the test accuracy (top) and test loss (bottom) of DNN and CNN models using 100\% of retained PCA variance across 30 executions.}
\label{fig:boxplot_results}
\end{figure}

To provide a comprehensive overview of model stability and variability using the full dataset, each experiment was repeated 30 times using the entire training set, with 100\% of the original variance retained (with all PCA dimensions). Figure~\ref{fig:boxplot_results} presents boxplots of test accuracy (top) and test loss (bottom) for both the DNN and CNN models, where the points are the individual executions. These plots illustrate the distribution of performance across all executions, highlighting aspects such as spread, central tendency, and the presence of outliers. The visual evidence reinforces the importance of multiple runs to account for variability due to random initialization and data shuffling.

To complement the visual analysis, a Kolmogorov–Smirnov (KS) test \cite{an1933sulla,smirnov1948table} was conducted to statistically compare the \textbf{test accuracy distributions} of the DNN and CNN models across 30 executions, using a 95\% confidence level. The test yielded a KS statistic of 0.3333 and a $p$-value of 0.0708, indicating no significant difference between the accuracy distributions of the two models.

In contrast, applying the KS test to the \textbf{test loss distributions} under the same confidence level (95\%) resulted in a KS statistic of 0.7000 and a $p$-value below 0.0001, evidencing a statistical difference. Although the CNN showed lower error values than the DNN, both are expected to achieve statistically equivalent performance in classification (test accuracy). Nevertheless, given this statistical equivalence in accuracy, the continuation of this work presents further analyses, including the best- and worst-case scenarios for each model.

\subsection{Average Performance Using the Full Dataset}

To complement the performance distribution shown previously, we now analyze the average behavior of both models across the 30 independent runs using the full training dataset, with 100\% of the original variance retained. This broader evaluation provides insights into each model’s stability, generalization capability, and sensitivity to initialization. The 95\% confidence intervals were calculated using Student’s t-test.

The DNN model achieved an average test accuracy of \textbf{91.37\%}, with a standard deviation of \textbf{7.41}, a median of \textbf{94.41\%}, and a confidence interval of \textbf{[88.59, 94.13]}. These results indicate that while the DNN is capable of high accuracy, its performance varies depending on initialization and training dynamics.

Under the same conditions, the CNN model achieved a slightly higher average test accuracy of \textbf{92.70\%}, with a standard deviation of \textbf{7.53}, a median of \textbf{95.00\%}, and a confidence interval of \textbf{[89.89, 95.51]}. This suggests that the CNN exhibits marginally better generalization, though its variability, reflected by the standard deviation, remains comparable to that of the DNN.

It is worth noting that the best and worst accuracies observed across the 30 runs vary considerably for both models. Specifically, the DNN achieved the highest accuracy of \textbf{95.67\%} and the lowest accuracy of \textbf{59.33\%}, while the CNN's best and worst accuracies were \textbf{97.67\%} and \textbf{55.83\%}, respectively. These extremes highlight the variability inherent in training deep learning models due to factors such as initialization and stochastic optimization.

A summary of these statistics is presented in Table~\ref{tab:avg_performance}, which includes the average test accuracy, standard deviation, and median values for both models. 

\begin{table}[!ht]
\centering
\caption{Average, standard deviation, and median test accuracy over 30 runs using 100\% of the data.}
\begin{tabular}{|l|c|c|c|}
\hline
\textbf{Model} & \textbf{Average (\%)} & \textbf{Std (\%)} & \textbf{Median (\%)} \\
\hline
DNN & 91.37 & 7.41 & 94.41 \\
CNN & 92.70 & 7.53 & 95.00 \\
\hline
\end{tabular}
\label{tab:avg_performance}
\end{table}

\begin{table}[!ht]
\centering
\caption{Best, worst, and 95\% confidence interval of test accuracy for DNN and CNN.}
\begin{tabular}{|l|c|c|c|}
\hline
\textbf{Model} & \textbf{Best (\%)} & \textbf{Worst (\%)} & \textbf{CI (95\%)} \\
\hline
DNN & 95.67 & 59.33 & [88.59, 94.13] \\
CNN & 97.67 & 55.83 & [89.89, 95.51] \\
\hline
\end{tabular}
\label{tab:best_worst}
\end{table}

Additionally, Table~\ref{tab:best_worst} highlights the best and worst performances observed during the 30 executions, along with the corresponding 95\% confidence intervals, illustrating the range of possible outcomes under different initializations.

\subsection{Data Proportion Impact on Model Performance}

The experimental results in this section were obtained by applying Principal Component Analysis (PCA) while retaining 95\%, 90\%, 85\%, and 80\% of the original variance. This dimensionality reduction preserves a controlled proportion of the dataset’s information, enabling the evaluation of model performance under different levels of data compression.

\begin{figure}[!ht]
\centerline{\includegraphics[width=\linewidth, height=13cm, keepaspectratio]{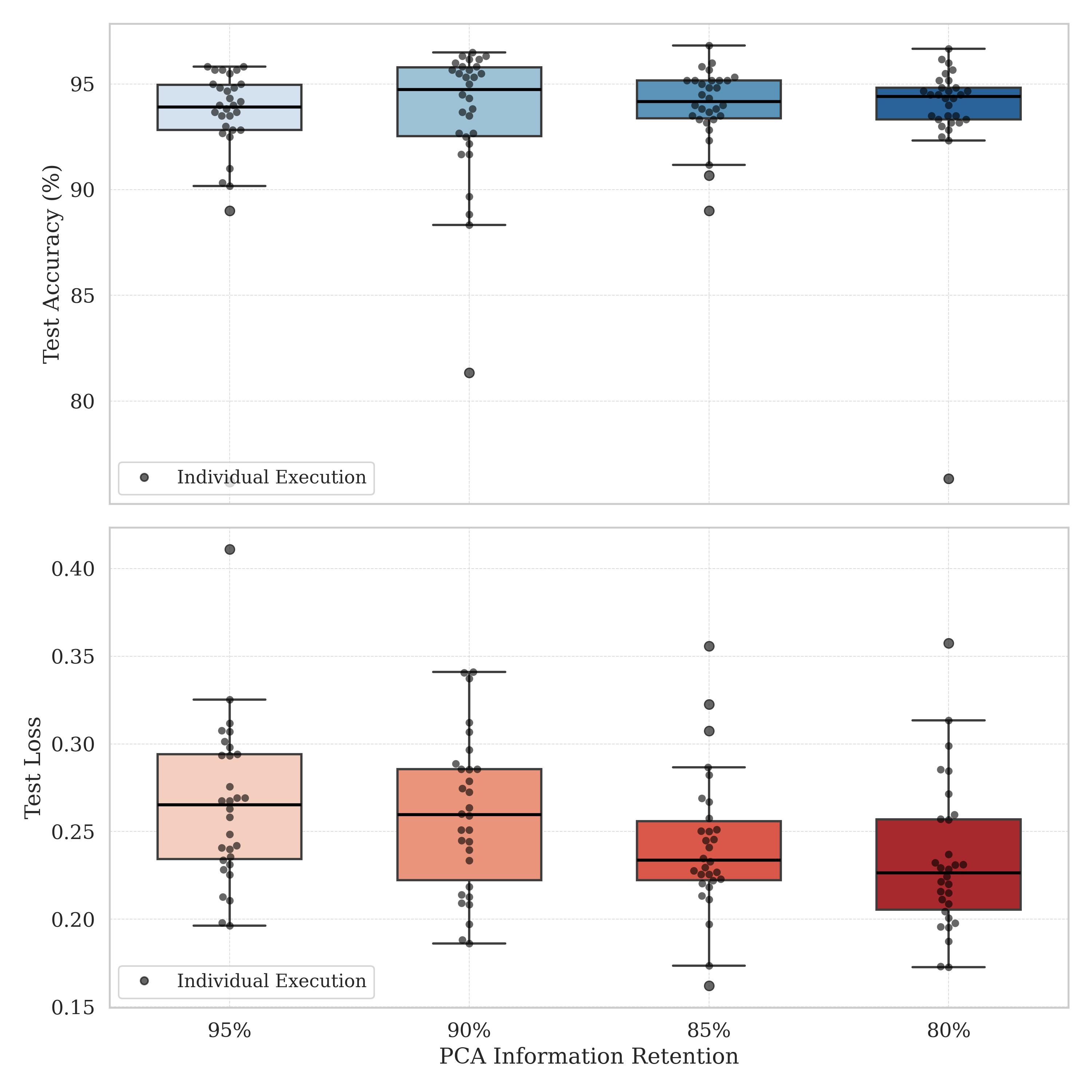}}
\caption{Boxplots showing the test accuracy (top) and test loss (bottom) of the DNN model under different PCA information retention levels (95\%, 90\%, 85\%, and 80\%) across 30 independent executions.}
\label{fig:boxplots_dnn_all}
\end{figure}

\begin{figure}[!ht]
\centerline{\includegraphics[width=\linewidth, height=13cm, keepaspectratio]{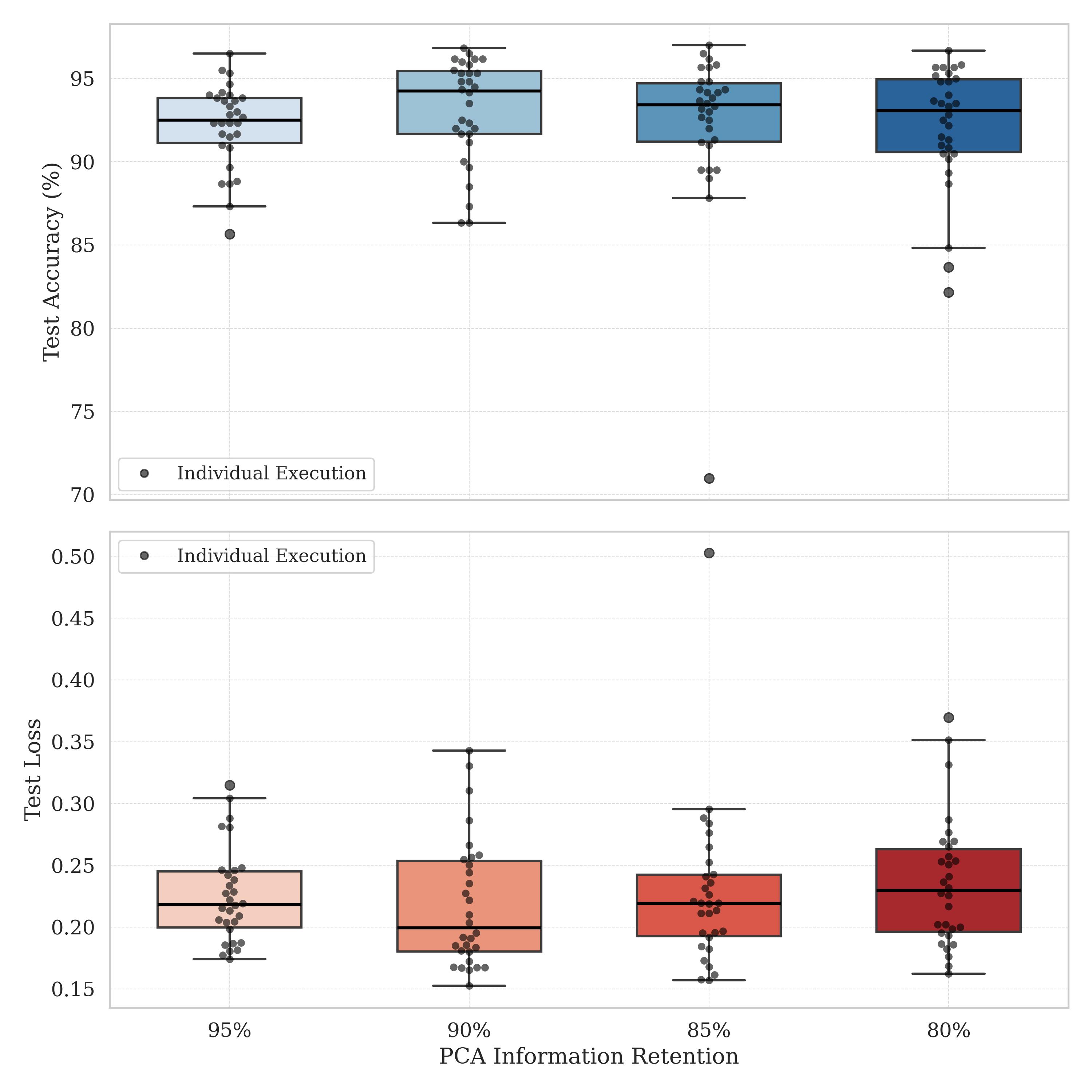}}
\caption{Boxplots showing the test accuracy (top) and test loss (bottom) of the CNN model under different PCA information retention levels (95\%, 90\%, 85\%, and 80\%) across 30 independent executions.}
\label{fig:boxplots_cnn_all}
\end{figure}

For each proportion, boxplots of test accuracy and loss across 30 executions are presented for both DNN and CNN models. This repeated evaluation accounts for the randomness inherent to neural network training and provides a robust estimate of generalization and stability. Figures~\ref{fig:boxplots_dnn_all} and~\ref{fig:boxplots_cnn_all} summarize these results, combining all four proportions of retained variance to allow direct comparison of the models. The boxplots reveal trends in variability and performance, showing how dimensionality reduction impacts robustness and consistency across executions.

\subsection{Statistical Summary of Model Performance}

A detailed statistical evaluation was conducted to assess the average behavior and consistency of both DNN and CNN models under different levels of data compression, specifically with 95\%, 90\%, 85\%, and 80\% of retained PCA variance. Tables~\ref{tab:accuracy_dnn} to~\ref{tab:std_deviation_cnn} present key performance metrics, including mean test accuracy, standard deviation, median, best and worst outcomes across 30 executions, and 95\% confidence intervals calculated using Student's t-test, providing a comprehensive overview of each model's performance variability and allowing a direct comparison of stability under varying amounts of retained information.

Tables~\ref{tab:accuracy_dnn} and~\ref{tab:accuracy_cnn} summarize the mean, standard deviation, and median of the test accuracy for each model across all training proportions. The DNN achieved its highest average accuracy (94.04\%) with the 85\% configuration, which also presented the lowest standard deviation (1.66\%) and a median of 94.16\%, indicating stable and consistent performance across runs and limited fluctuation between executions. Similarly, the CNN reached its peak average accuracy of 93.09\% with the 90\% configuration, accompanied by a median of 94.25\%, although the variability across runs was slightly higher compared to the DNN, as reflected by its standard deviation. These results highlight the relative stability of both models, the slight differences in sensitivity to PCA variance retention, and provide a clear picture of how the proportion of retained features affects predictive consistency and overall performance.

\begin{table}[!ht]
\centering
\caption{DNN model: mean, standard deviation, and median of test accuracy over 30 runs for different training proportions.}
\begin{tabular}{|l|c|c|c|}
\hline
\textbf{Proportion (\%)} & \textbf{Mean (\%)} & \textbf{Std (\%)} & \textbf{Median (\%)} \\
\hline
95  & 93.12 & 3.65 & 93.91 \\
90  & 93.62 & 3.25 & 94.75 \\
85  & 94.04 & 1.66 & 94.16 \\
80  & 93.69 & 3.45 & 94.41 \\
\hline
\end{tabular}
\label{tab:accuracy_dnn}
\end{table}

\begin{table}[!ht]
\centering
\caption{CNN model: mean, standard deviation, and median of test accuracy over 30 runs for different training proportions.}
\begin{tabular}{|l|c|c|c|}
\hline
\textbf{Proportion (\%)} & \textbf{Mean (\%)} & \textbf{Std (\%)} & \textbf{Median (\%)} \\
\hline
95  & 92.19 & 2.50 & 92.50 \\
90  & 93.09 & 3.10 & 94.25 \\
85  & 92.37 & 4.69 & 93.41 \\
80  & 92.16 & 3.63 & 93.08 \\
\hline
\end{tabular}
\label{tab:accuracy_cnn}
\end{table}

Tables~\ref{tab:std_deviation_dnn} and~\ref{tab:std_deviation_cnn} provide additional insight into model robustness by reporting the best and worst performances observed, along with the respective 95\% confidence intervals. The DNN displayed remarkable consistency at 85\% retained variance, with accuracy values ranging from 89.00\% to 96.83\% and a narrow confidence interval of [93.41\%, 94.65\%]. Meanwhile, the CNN achieved its best individual performance (97.00\%) at the same 85\% level, though it exhibited greater dispersion, with the lowest performance dropping to 71.00\%.

\begin{table}[!ht]
\centering
\caption{DNN model: best, worst, and 95\% confidence interval of test accuracy over 30 runs.}
\begin{tabular}{|l|c|c|c|}
\hline
\textbf{Proportion (\%)} & \textbf{Best (\%)} & \textbf{Worst (\%)} & \textbf{CI (95\%)} \\
\hline
95  & 95.83 & 76.17 & [91.75, 94.48] \\
90  & 96.50 & 81.33 & [92.40, 94.83] \\
85  & 96.83 & 89.00 & [93.41, 94.65] \\
80  & 96.67 & 76.33 & [92.39, 94.98] \\
\hline
\end{tabular}
\label{tab:std_deviation_dnn}
\end{table}

\begin{table}[!ht]
\centering
\caption{CNN model: best, worst, and 95\% confidence interval of test accuracy over 30 runs.}
\begin{tabular}{|l|c|c|c|}
\hline
\textbf{Proportion (\%)} & \textbf{Best (\%)} & \textbf{Worst (\%)} & \textbf{CI (95\%)} \\
\hline
95  & 96.50 & 87.33 & [91.26, 93.12] \\
90  & 96.83 & 86.33 & [91.93, 94.25] \\
85  & 97.00 & 71.00 & [90.61, 94.11] \\
80  & 96.67 & 82.17 & [90.79, 93.51] \\
\hline
\end{tabular}
\label{tab:std_deviation_cnn}
\end{table}

Despite fluctuations in the extreme values, both models maintained relatively narrow confidence intervals across all data proportions, reinforcing their ability to generalize well under varying degrees of information retention. These results suggest that moderate dimensionality reduction not only preserves classification performance but may also enhance model stability in some cases.

\subsection{Best-Case Evaluation of DNN and CNN Models}

To better illustrate the classification capabilities of both deep learning models, this subsection presents a detailed analysis of the best-performing run out of the 30 independent executions for each architecture. This includes training performance plots, confusion matrices, and classification reports, offering a comprehensive view of how each model behaves when operating at its full potential.

For the DNN model, the best result was obtained using PCA with 85\% of the original variance retained, achieving a test accuracy of 96.83\%. This shows that moderate dimensionality reduction preserved the most relevant features while potentially reducing noise. In contrast, the best CNN model did not employ dimensionality reduction, retaining 100\% of the features, and achieved a slightly higher test accuracy of 97.67\%, suggesting that this architecture benefits from the full feature set.

\subsubsection{DNN Best-Case Performance (PCA 85\%)}

\begin{figure}[!ht]
\centering
\includegraphics[width=\linewidth, height=6.2cm, keepaspectratio]{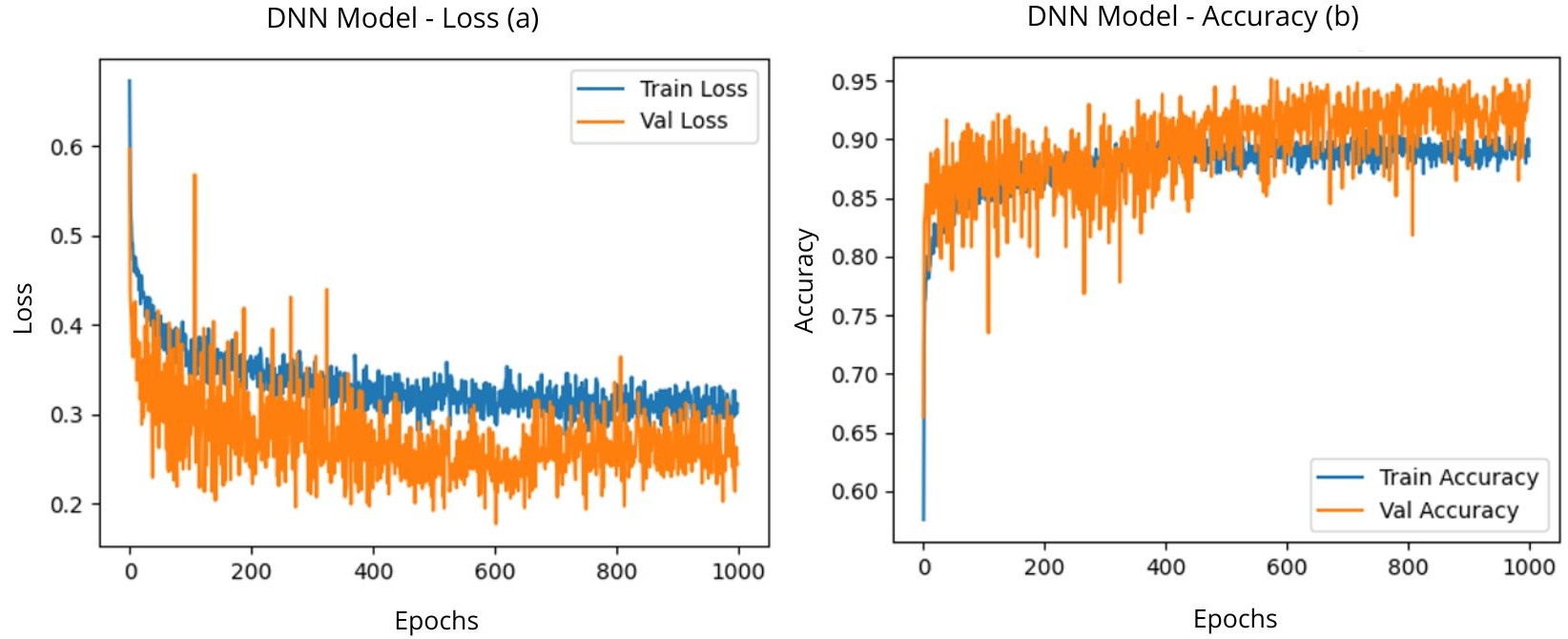}
\caption{Training and validation accuracy and loss of the best-performing DNN model using 85\% PCA variance.}
\label{fig:dnn_best_train_curve}
\end{figure}

Figure~\ref{fig:dnn_best_train_curve} presents the training behavior of the best DNN model. It contains two subplots: one showing training and validation accuracy, and the other showing training and validation loss across epochs. The model reached a final training accuracy of 88.61\% and a validation accuracy of 95.00\%, indicating effective generalization. The smooth convergence of both curves suggests stable training without overfitting.

Table~\ref{tab:dnn_best_confusion} shows the corresponding confusion matrix, revealing balanced performance across both classes. The model correctly classified most samples, with only a small number of misclassifications, 11 for class 0 and 8 for class 1, indicating strong overall predictive capability. The classification report for this execution is shown in Table~\ref{tab:dnn_best_classification}. The model achieved high precision, recall, and F1-scores for both classes, demonstrating balanced performance and strong predictive capability. Both macro and weighted averages are 0.97, reflecting consistent accuracy across the dataset. The support values indicate that the evaluation included 323 samples of Undefined Loop (0) and 277 samples of Independent Loop (1), confirming that the model performed reliably across classes of differing sizes. Overall, these results highlight the DNN's robustness in correctly classifying the majority of samples and maintaining high performance metrics throughout.

\begin{table}[!ht]
\centering
\caption{Confusion matrix of the best-performing DNN model.}
\begin{tabular}{|c|c|c|}
\hline
 & Predicted 0 & Predicted 1 \\
\hline
Actual 0 & 312 & 11 \\
Actual 1 & 8   & 269 \\
\hline
\end{tabular}
\label{tab:dnn_best_confusion}
\end{table}

\begin{table}[!ht]
\centering
\caption{Classification report of the best-performing DNN model.}
\begin{tabular}{|l|c|c|c|c|}
\hline
Class & Precision & Recall & F1-score & Support \\
\hline
Undefined Loop (0) & 0.97 & 0.97 & 0.97 & 323 \\
Independent Loop (1) & 0.96 & 0.97 & 0.97 & 277 \\
\hline
\textbf{Accuracy} & \multicolumn{4}{c|}{\textbf{0.97 (96.83\%)}} \\
\hline
Macro avg & 0.97 & 0.97 & 0.97 & 600 \\
Weighted avg & 0.97 & 0.97 & 0.97 & 600 \\
\hline
\end{tabular}
\label{tab:dnn_best_classification}
\end{table}

\subsubsection{CNN Best-Case Performance (Full Features)}

\begin{figure}[!ht]
\centering
\includegraphics[width=\linewidth, height=6.2cm, keepaspectratio]{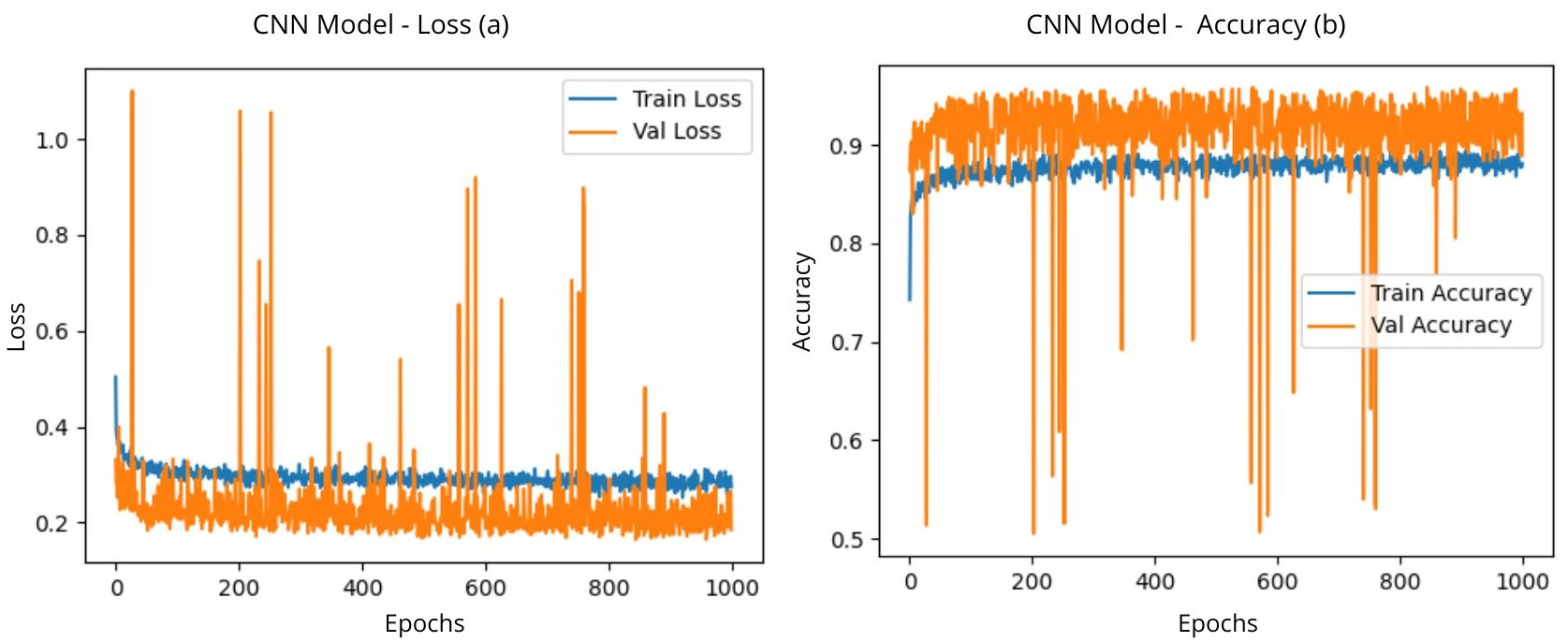}
\caption{Training and validation accuracy and loss of the best-performing CNN model using 100\% of the features.}
\label{fig:cnn_best_train_curve}
\end{figure}

Figure~\ref{fig:cnn_best_train_curve} shows the training and validation performance of the best CNN model. Similar to the DNN, it includes two subplots: one displaying accuracy curves and the other showing loss curves. The CNN reached a training accuracy of 88.11\% and a validation accuracy of 93.17\%, indicating effective learning of the training data while generalizing well to unseen samples. The convergence of both curves and the relatively small gap between training and validation metrics suggest stable training, consistent improvement over epochs, and the absence of overfitting. 

The confusion matrix in Table~\ref{tab:cnn_best_confusion} shows excellent classification performance, with zero false positives for class 0 and a small number of false negatives in class 1.

\begin{table}[!ht]
\centering
\caption{Confusion matrix of the best-performing CNN model.}
\begin{tabular}{|c|c|c|}
\hline
 & Predicted 0 & Predicted 1 \\
\hline
Actual 0 & 323 & 0 \\
Actual 1 & 14  & 263 \\
\hline
\end{tabular}
\label{tab:cnn_best_confusion}
\end{table}

The classification report for this execution, presented in Table~\ref{tab:cnn_best_classification}, confirms a slightly higher overall accuracy compared to the DNN. Both macro and weighted averages reach 0.98, demonstrating consistent and balanced performance across classes. The high recall for class 0 and strong precision for class 1 emphasize the model’s ability to correctly identify each category while minimizing misclassifications.

\begin{table}[!ht]
\centering
\caption{Classification report of the best-performing CNN model.}
\begin{tabular}{|l|c|c|c|c|}
\hline
Class & Precision & Recall & F1-score & Support \\
\hline
Undefined Loop (0) & 0.96 & 1.00 & 0.98 & 323 \\
Independent Loop (1) & 1.00 & 0.95 & 0.97 & 277 \\
\hline
\textbf{Accuracy} & \multicolumn{4}{c|}{\textbf{0.98 (97.67\%)}} \\
\hline
Macro avg & 0.98 & 0.97 & 0.98 & 600 \\
Weighted avg & 0.98 & 0.98 & 0.98 & 600 \\
\hline
\end{tabular}
\label{tab:cnn_best_classification}
\end{table}

\subsection{Worst-Case Evaluation of DNN and CNN Models}

While the previous subsection highlighted the best-performing runs, this section focuses on the opposite end of the performance spectrum. Here, we examine the worst-performing runs for both the DNN and CNN models. As in the previous analysis, the evaluation includes training plots, confusion matrices, and classification reports, allowing for a detailed assessment of the models. By analyzing the worst-case performance, we can gain a deeper understanding of the limitations of each model and the scenarios in which they fail to generalize effectively.

\subsubsection{DNN Worst-Case Performance (Full Features)}

\begin{figure}[!ht]
\centering
\includegraphics[width=\linewidth, height=6.2cm, keepaspectratio]{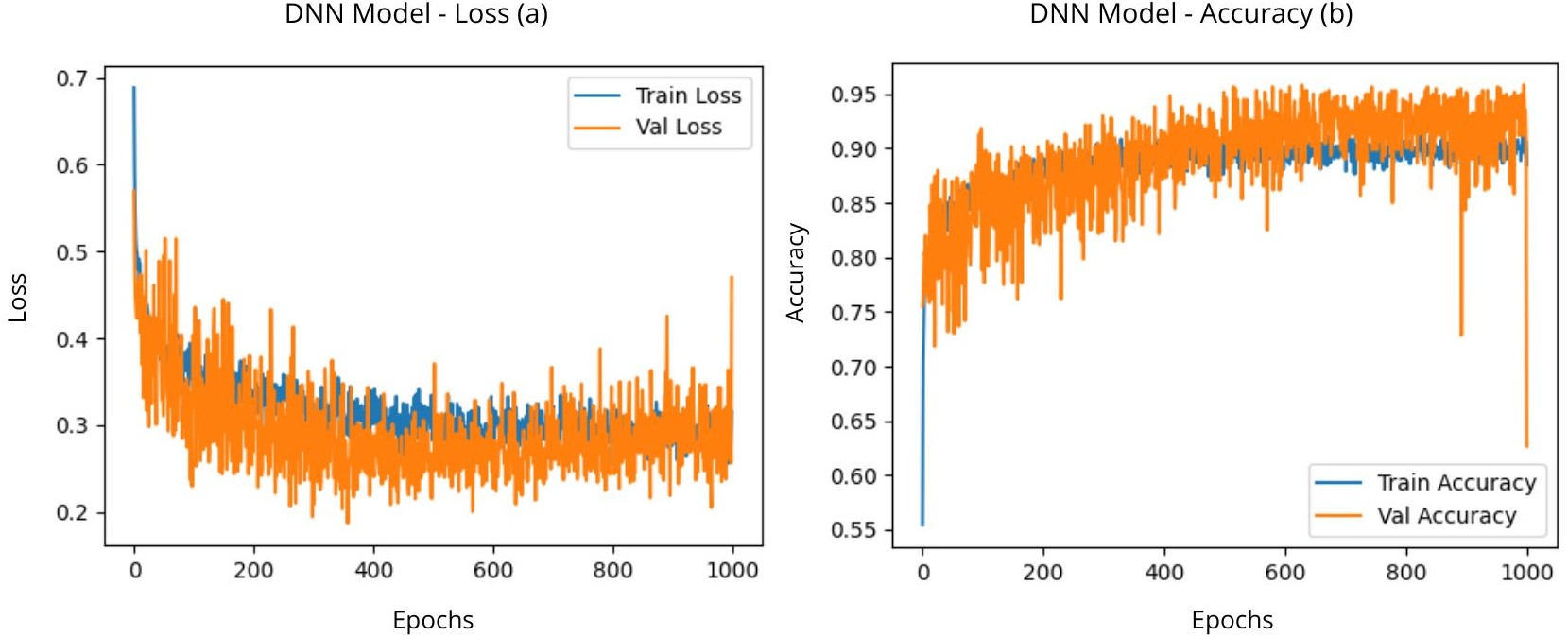}
\caption{Training and validation accuracy and loss of the worst-performing DNN model using 100\% of the features.}
\label{fig:dnn_worst_train_curve}
\end{figure}

Figure~\ref{fig:dnn_worst_train_curve} shows in detail the training behavior of the worst-performing DNN model, which was configured using 100\% of the original features as input. The figure is divided into two subplots: the first illustrates the evolution of training and validation accuracy over the epochs, and the second presents the curves of training and validation loss. At the end of training, the model reached a relatively high training accuracy of 88.50\%, but its validation accuracy dropped considerably to 62.67\%. This sharp decline clearly reflects strong overfitting, since the model demonstrated good performance on the training set but failed to reproduce similar results on the validation set. The large gap between training and validation curves shows that the network tended to memorize the training data instead of learning patterns capable of generalizing to new, unseen samples. This discrepancy highlights the sensitivity of the DNN model to the specific characteristics of the training data and reinforces the importance of simultaneously analyzing both training and validation metrics when assessing the overall reliability and robustness of the model.

Table~\ref{tab:dnn_worst_confusion} presents the confusion matrix of the worst-performing DNN model, revealing a strong class imbalance in prediction. The model failed to correctly classify most instances of class 0 (Undefined Loop), with 244 misclassifications, while class 1 (Independent Loop) was classified correctly in all cases. This imbalance is further highlighted in the classification report shown in Table~\ref{tab:dnn_worst_classification}, where the model achieved perfect recall for class 1 but very low recall for class 0, resulting in a low F1-score for that class and an overall accuracy of 59.33\%.

\begin{table}[!ht]
\centering
\caption{Confusion matrix of the worst-performing DNN model.}
\begin{tabular}{|c|c|c|}
\hline
 & Predicted 0 & Predicted 1 \\
\hline
Actual 0 & 79  & 244 \\
Actual 1 & 0   & 277 \\
\hline
\end{tabular}
\label{tab:dnn_worst_confusion}
\end{table}

\begin{table}[!ht]
\centering
\caption{Classification report of the worst-performing DNN model.}
\begin{tabular}{|l|c|c|c|c|}
\hline
Class & Precision & Recall & F1-score & Support \\
\hline
Undefined Loop (0) & 1.00 & 0.24 & 0.39 & 323 \\
Independent Loop (1) & 0.53 & 1.00 & 0.69 & 277 \\
\hline
\textbf{Accuracy} & \multicolumn{4}{c|}{\textbf{0.59 (59.33\%)}} \\
\hline
Macro avg & 0.77 & 0.62 & 0.54 & 600 \\
Weighted avg & 0.78 & 0.59 & 0.53 & 600 \\
\hline
\end{tabular}
\label{tab:dnn_worst_classification}
\end{table}

\subsubsection{CNN Worst-Case Performance (Full Features)}

\begin{figure}[!ht]
\centering
\includegraphics[width=\linewidth, height=6.2cm, keepaspectratio]{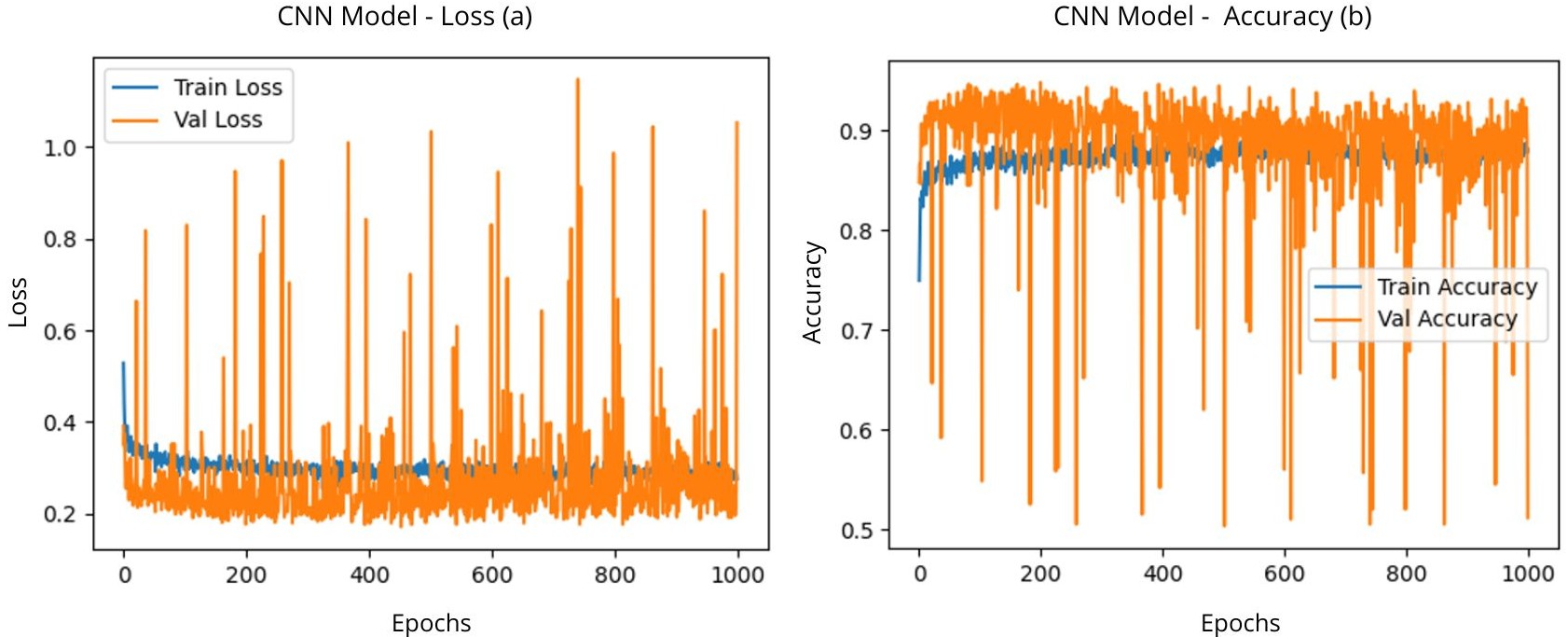}
\caption{Training and validation accuracy and loss of the worst-performing CNN model using 100\% of the features.}
\label{fig:cnn_worst_train_curve}
\end{figure}

The worst-performing CNN model also used the full feature set (100\% variance retained). Its training curves are shown in Figure~\ref{fig:cnn_worst_train_curve}, which includes accuracy and loss plots for both training and validation. The final training accuracy was 88.14\%, while validation accuracy dropped to 51.17\%, again suggesting overfitting.

As shown in Table~\ref{tab:cnn_worst_confusion}, the confusion matrix reveals a highly imbalanced prediction pattern. The model correctly identified nearly all instances of class 0 (Undefined Loop), but severely misclassified most of class 1 (Independent Loop), with only 13 correct predictions.

\begin{table}[!ht]
\centering
\caption{Confusion matrix of the worst-performing CNN model.}
\begin{tabular}{|c|c|c|}
\hline
 & Predicted 0 & Predicted 1 \\
\hline
Actual 0 & 322 & 1 \\
Actual 1 & 264 & 13 \\
\hline
\end{tabular}
\label{tab:cnn_worst_confusion}
\end{table}

The classification report in Table~\ref{tab:cnn_worst_classification} confirms the poor generalization of this model, particularly for class 1 (Independent Loop). While it reached a high precision of 0.93 for this class, the recall dropped drastically to 0.05, and the F1-score was only 0.09, indicating that the model failed to correctly identify most instances of this class. In contrast, class 0 (Undefined Loop) achieved perfect recall (1.00) but low precision (0.55), showing that many predictions were misclassified as this class. Overall, these imbalances resulted in a low overall accuracy of 55.83\%, with macro and weighted averages also reflecting poor performance (0.40 and 0.42 F1-score, respectively). This highlights the model’s inability to handle class-specific distinctions effectively under worst-case conditions.

\begin{table}[!ht]
\centering
\caption{Classification report of the worst-performing CNN model.}
\begin{tabular}{|l|c|c|c|c|}
\hline
Class & Precision & Recall & F1-score & Support \\
\hline
Undefined Loop (0) & 0.55 & 1.00 & 0.71 & 323 \\
Independent Loop (1) & 0.93 & 0.05 & 0.09 & 277 \\
\hline
\textbf{Accuracy} & \multicolumn{4}{c|}{\textbf{0.56 (55.83\%)}} \\
\hline
Macro avg & 0.74 & 0.52 & 0.40 & 600 \\
Weighted avg & 0.72 & 0.56 & 0.42 & 600 \\
\hline
\end{tabular}
\label{tab:cnn_worst_classification}
\end{table}

These results highlight the variability that can arise across different executions of the same model architecture. While both DNN and CNN achieved excellent performance in their best runs, their worst cases revealed significant drops in accuracy and class imbalance issues. This underscores the importance of evaluating model robustness through multiple runs, as individual executions may lead to widely different outcomes depending on initialization and training dynamics.

\section{Conclusion}
\label{sec:conclusions}

This study has demonstrated the effectiveness of deep learning approaches for classifying programming code based on its parallelization potential. Using a carefully curated dataset of independent and undefined loops, we implemented and evaluated both DNN and CNN architectures. Both models achieved strong average performance across 30 runs and showed robust generalization across training proportions (80–100\%), with comparable variability, as indicated by their similar standard deviations. To further support these findings, a Kolmogorov–Smirnov (KS) test was applied to the accuracy and loss distributions of both models. The results revealed no statistically significant difference in test accuracy at the 95\% confidence level, while the test loss distributions were statistically different, with the CNN achieving lower error values. These results indicate that, although both models are statistically equivalent in terms of classification accuracy, the CNN demonstrated slightly superior performance in its best-case scenario.

The findings carry important implications for automated code analysis tools, particularly in the domain of parallel computing. The CNN's consistent performance across multiple executions suggests that convolutional operations are especially effective at capturing the structural regularities that distinguish parallelizable loops. This work lays the groundwork for assessing parallelization potential using deep learning techniques. While existing literature on parallelization methods—including those cited in this study—provides useful context, our approach diverges by focusing specifically on classifying loops based on learned structural patterns rather than handcrafted heuristic rules and other parallelization identification methods. This methodological distinction limited the possibility of direct comparisons with prior techniques within our experimental setup.

Looking ahead, several promising directions emerge for extending this research. Expanding the dataset to include more diverse real-world code samples would further enhance model generalization. Testing alternative architectures, such as transformer-based models or hybrid CNN-RNN approaches, could provide additional insights into optimal network designs for this task. Practical validation through application to open-source projects would assess the real-world utility of the approach. Most significantly, future work should focus on formalizing mathematical relationships that determine loop independence, potentially leading to the definition of a new code smell category specifically for parallelization analysis. These advancements would build upon the robust framework established in this study, which demonstrates both the feasibility and potential of deep learning for automated parallelization assessment.


\bibliographystyle{plainnat}
\bibliography{BibTex-Sample}

\end{document}